\documentclass[runningheads]{llncs}
\usepackage{graphicx}

\usepackage{tikz}
\usepackage{comment}
\usepackage{amsmath,amssymb} 
\usepackage{color}

\usepackage{multirow}
\usepackage{bbding}
\usepackage{makecell}
\usepackage{amsfonts}
\usepackage{comment}
\usepackage{xcolor}
\usepackage{epstopdf}
\usepackage{comment}
\usepackage{subfigure}
\usepackage{booktabs}

\definecolor{mygray}{gray}{.9}
\usepackage{colortbl}

\usepackage[accsupp]{axessibility}  

\begin{document}
\pagestyle{headings}
\mainmatter
\def\ECCVSubNumber{100}  

\title{Few-Shot Object Detection by Knowledge Distillation Using Bag-of-Visual-Words Representations} 

\titlerunning{Knowledge Distillation Using Bag-of-Visual-Words Representations}
%
\author{Wenjie Pei\inst{2,\dagger} \and Shuang Wu\inst{2,\dagger} \and   Dianwen Mei\inst{2} \and Fanglin Chen\inst{2} \and Jiandong Tian\inst{3} \and Guangming Lu\inst{1,2,*}}
\authorrunning{W. Pei and S. Wu et al.}
%
\institute{Guangdong Provincial Key Laboratory of Novel Security Intelligence Technologies \\ \and Harbin Institute of Technology, Shenzhen, China \\
\and Shenyang Institute of Automation, Chinese Academy of Sciences \\
\email{\{wenjiecoder, wushuang9811\}@outlook.com},
\email{\{178mdw, linwers\}@gmail.com},
\email{luguangm@hit.edu.cn},
\email{tianjd@sia.cn}
}

\maketitle

\renewcommand{\thefootnote}{}
\footnotetext{$^\dagger$ Equal contribution. \\ $^*$ Corresponding author.}

\begin{abstract}
While fine-tuning based methods for few-shot object detection have achieved remarkable progress, a crucial challenge that has not been addressed well is the potential class-specific overfitting on base classes and sample-specific overfitting on novel classes. In this work we design a novel knowledge distillation framework to guide the learning of the object detector and thereby restrain the overfitting in both the pre-training stage on base classes and fine-tuning stage on novel classes. To be specific, we first present a novel Position-Aware Bag-of-Visual-Words model for learning a representative bag of visual words (\emph{BoVW}) from a limited size of image set, which is used to encode general images based on the similarities between the learned visual words and an image. Then we perform knowledge distillation based on the fact that an image should have consistent \emph{BoVW} representations in two different feature spaces. To this end, we pre-learn a feature space independently from the object detection, and encode images using \emph{BoVW} in this space. The obtained \emph{BoVW} representation for an image can be considered as distilled knowledge to guide the learning of object detector: the extracted features by the object detector for the same image are expected to derive the consistent \emph{BoVW} representations with the distilled knowledge. Extensive experiments validate the effectiveness of our method and demonstrate the superiority over other state-of-the-art methods.
\end{abstract}

\section{Introduction}
\label{sec:intro}
Few-shot object detection aims to learn effective object detectors on a set of base classes with sufficient samples, which can be generalized efficiently to novel classes with only a few samples available. Thus, few-shot object detection eliminates exhaustive label annotation of massive data on novel classes. Compared to general object detection~\cite{lin2017focal,redmon2017yolo9000,ren2015faster,tian2019fcos}, few-shot object detection~\cite{chen2018lstd,kang2019few} is much more challenging due to the difficulty of learning generalizable features that can be transferred from base classes to novel classes.  

\begin{figure}[t]
  \centering
   \includegraphics[width=0.65\linewidth]{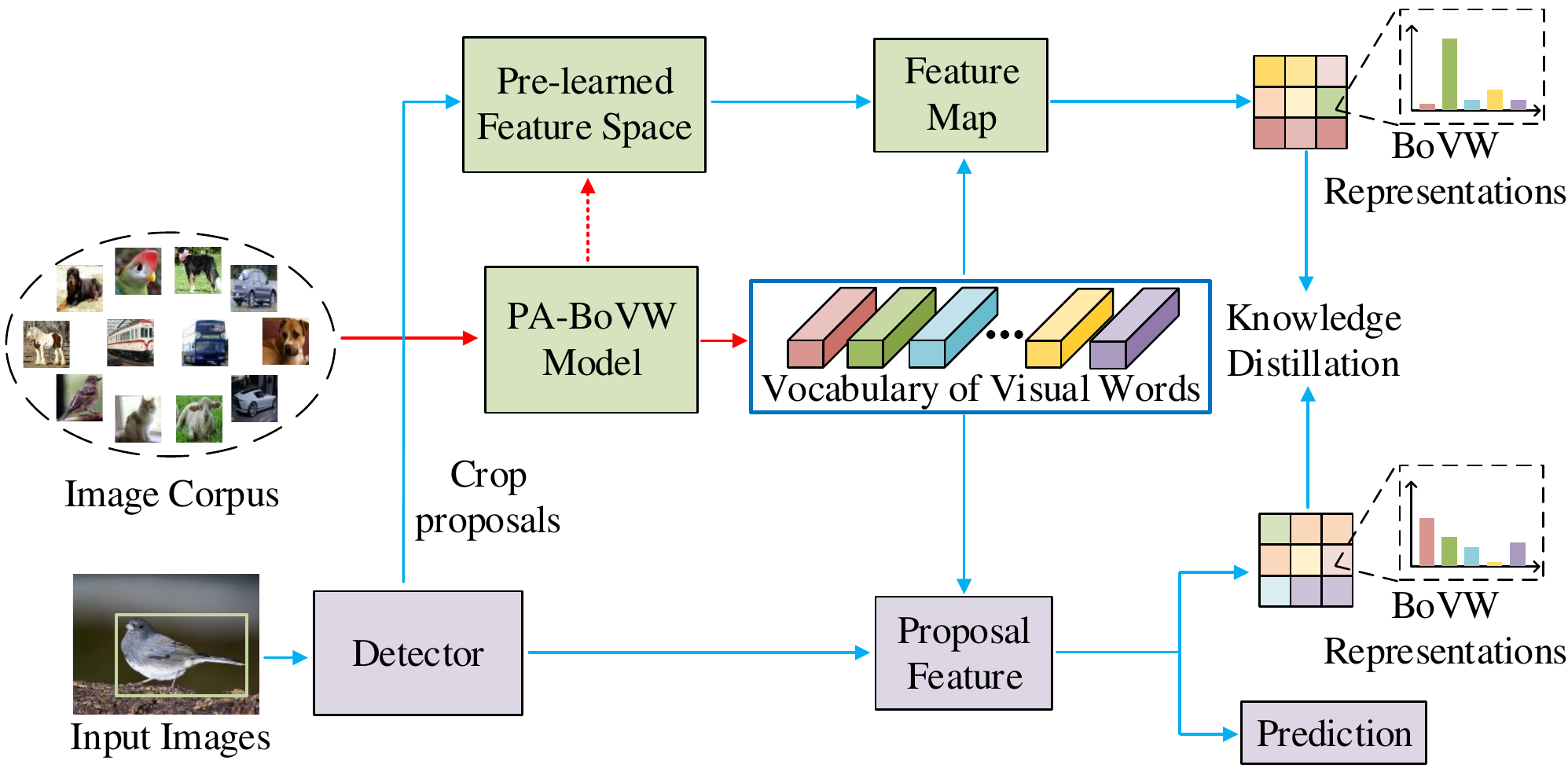}
   \caption{We learn a representative bag of visual words (\emph{BoVW}) using the proposed \emph{PA-BoVW} model. For an extracted positive proposal, we encode it with \emph{BoVW} in a pre-learned feature space and the feature space of the object detector, respectively. Then we perform knowledge distillation by matching two \emph{BoVW} representations to guide the learning of the object detector.}
   \label{fig:intro}
\end{figure}

A classical type of methods for few-shot object detection is fine-tuning based methods~\cite{fadi2021,fan2021generalized,li2021few,qiao2021defrcn,sun2021fsce,wang2020frustratingly,svd2021,wu2020multi,zhang2021hallucination}, which first train the object detector using the samples from base classes, then fine-tune the model on novel classes. A prominent example is TFA~\cite{wang2020frustratingly}, which first adopts such two-stage training strategy to transfer knowledge from base classes to novel classes. Based on such fine-tuning framework, many methods are proposed to deal with various challenges of few-shot object detection. Typical methods include FSCE~\cite{sun2021fsce} aiming to facilitate the separability among similar classes, MPSR~\cite{wu2020multi} which seeks to rectify the sample distribution for novel classes, and HallucFsDet~\cite{zhang2021hallucination} which is designed to tackle the problem of data scarcity.

A crucial challenge of fine-tuning based framework for few-shot object detection is the potential class-specific overfitting on base classes and sample-specific overfitting on novel classes. On the one hand, although sufficient samples are provided for base classes, the object detector is still prone to overfitting on base classes during the first stage of training process. In this case, the detector learns the class-specific features instead of the class-agnostic features, which cannot be transferred to novel classes and would adversely affect object detection for novel classes. On the other hand, owing to the scarcity of training samples for novel classes in the fine-tuning stage, the object detector tends to be overfitting on these individual samples and thus learns sample-specific features that cannot be generalized across different samples for a same novel class.

To address above limitation, we propose to perform knowledge distillation to guide the learning process of few-shot object detection and thus restrain the potential overfitting on both base classes and novel classes. As shown in Figure~\ref{fig:intro}, we propose the novel Position-Aware Bag-of-Visual-Words (\emph{PA-BoVW}) model, which is able to learn a bag of visual words (\emph{BoVW}) from a limited size of image set. The learned visual words are representative and comprehensive to be capable of encoding general images based on the similarities between the learned visual words and an image. Then we can perform knowledge distillation based on the intuition that an image should have consistent \emph{BoVW} representations in two different feature spaces, provided that the image is encoded properly, namely not overfitted, in both feature spaces. Concretely, we first pre-learn a feature space and derive a \emph{BoVW} representation for an image in this space. The obtained \emph{BoVW} representation can be considered as distilled knowledge to guide the learning of object detector: the extracted features by the object detector for the same image are expected to derive the consistent \emph{BoVW} representation with the distilled knowledge.

Unlike typical way that identifies visual words as the clustering centroids in the deep feature space~\cite{gidaris2020learning,jain2020quest}, we learn visual words as learnable vectorial embeddings. To be specific, our proposed \emph{PA-BoVW} model first constructs an effective deep embedding space for learning the visual words by training a backbone network in a self-supervised way. Then the visual words is learned in this embedding space in a supervised way employing image classification as a pretext task. Besides, we employ DeCov loss~\cite{cogswell2015reducing} as an auxiliary loss to reduce the inter-word redundancy and encourage the diversity of visual words. As a result, the \emph{PA-BoVW} model is learned in an independent way from the task of object detection. Thus the encoded \emph{BoVW} representation in its embedding space can be used as distilled knowledge, which can be transferred to the learning process of the detector to restrain potential overfitting on both base and novel classes. 

To conclude, we make following contributions: 1) We propose the novel \emph{PA-BoVW} model, which constructs an effective embedding space to learn a representative vocabulary of visual words; 2) Based on the \emph{PA-BoVW} model, we design a knowledge distillation framework to guide the learning of the object detector and thereby restrain the potential overfitting on both base classes and novel classes; 3) Extensive experiments validate the effectiveness of our method and demonstrate the advantages of our method over state-of-the-art methods for few-shot object detection.

\section{Related Work}
\label{sec:relatedwork}
\noindent\textbf{Few-shot learning.} Early works of few-shot learning focus on the task of image classification. Metric-based methods learn a suitable embedding space, where samples can be categorized correctly via a nearest neighbor classifier with Euclidean distance~\cite{snell2017prototypical}, cosine similarity~\cite{chen2018closer,vinyals2016matching} or graph distance~\cite{kim2019edge,satorras2018few,yang2020dpgn}. Initialization-based methods aim to learn good initialization so that the model can adapt to novel tasks by a few optimization steps~\cite{finn2017model,lee2019meta}.  Hallucination-based methods alleviate data scarcity issue via learning generators to augment novel classes~\cite{hariharan2017low,wang2018low}. However, these approaches could not be directly applied to few-shot object detection which requires both classification and localization.

\noindent\textbf{Few-shot object detection.} Few-shot object detection aims to detect objects with few annotated training examples provided. There are several early methods adopting the idea of meta-learning~\cite{fan2020few,han2021query,hu2021dense,kang2019few,li2021transformation,xiao2020few,yan2019meta,zhang2021accurate}. FSRW~\cite{kang2019few} is a novel few-shot detector based on YOLOv2~\cite{redmon2017yolo9000}, which re-weights the features with channel-wise attention and leverages these features to detect novel objects. Meta R-CNN~\cite{yan2019meta} applies similar feature re-weighting scheme to Faster R-CNN~\cite{ren2015faster} and performs meta-learning over RoI features. These methods usually suffer from a complicated training process and fail to learn generalizable features that can be transferred from base classes to novel classes. Recently, several fine-tuning based methods~\cite{fadi2021,fan2021generalized,li2021few,qiao2021defrcn,sun2021fsce,wang2020frustratingly,svd2021,wu2020multi,zhang2021hallucination} achieve higher performance compared to meta-learning based methods. TFA~\cite{wang2020frustratingly} performs a simple two-stage fine-tuning approach which fine-tunes only the last layer on novel classes. MPSR~\cite{wu2020multi} proposes to generate multi-scale positive samples to solve the problem of scale variations. FSCE~\cite{sun2021fsce} provides a strong baseline which fine-tunes feature extractors during the fine-tuning stage and employs a contrastive branch to rescue misclassifications. However, all these fine-tuning based methods suffer from overfitting on both base classes and novel classes. In this work, we design a novel knowledge distillation framework to tackle the problem.

\noindent\textbf{Bag of visual words.} Bag-of-Visual-Words is a popular technique for image recognition. Many variants of \emph{BoVW} have been proposed in the past\cite{csurka2004visual,lazebnik2006beyond,wang2010locality} and they continue to be widely used in recent deep learning approaches\cite{gidaris2020learning,gidaris2021obow,jain2020quest,ru2022weakly}. VWE\cite{ru2022weakly} designs a visual words learning module to generate CAMs\cite{zhou2016learning} for weakly-supervised semantic segmentation. BoWNet\cite{gidaris2020learning} and OBOW\cite{gidaris2021obow} apply \emph{BoVW} to self-supervised learning. QuEST\cite{jain2020quest} introduces to distill the quantized feature maps from the teacher to the student. In this work, we learn a \emph{BoVW} model via a self-supervised task and image classification task.

\noindent\textbf{Knowledge distillation for object detection.} Knowledge distillation~\cite{hinton2015distilling} is an effective way to transfer knowledge acquired in teacher network to student network. Early works focus on the task of image classification~\cite{romero2014fitnets,tian2019contrastive,yim2017gift}. Recently, there are several works which propose to transfer knowledge for object detection. Chen et al.~\cite{chen2017learning} distill knowledge from the teacher detector to the student detector in all components (i.e., feature extraction, RPN, classification and regression networks). Wang et al.~\cite{wang2019distilling} design a fine-grained feature imitation method which distills the features from foreground area. In this work, we propose a novel knowledge distillation method which transfers knowledge from a \emph{BoVW} model to a few-shot object detector, aiming to suppress potential overfitting on both base and novel classes.

\section{Method}
\label{sec:method}
\begin{figure}[!t]
  \centering
   \includegraphics[width=1.0\linewidth]{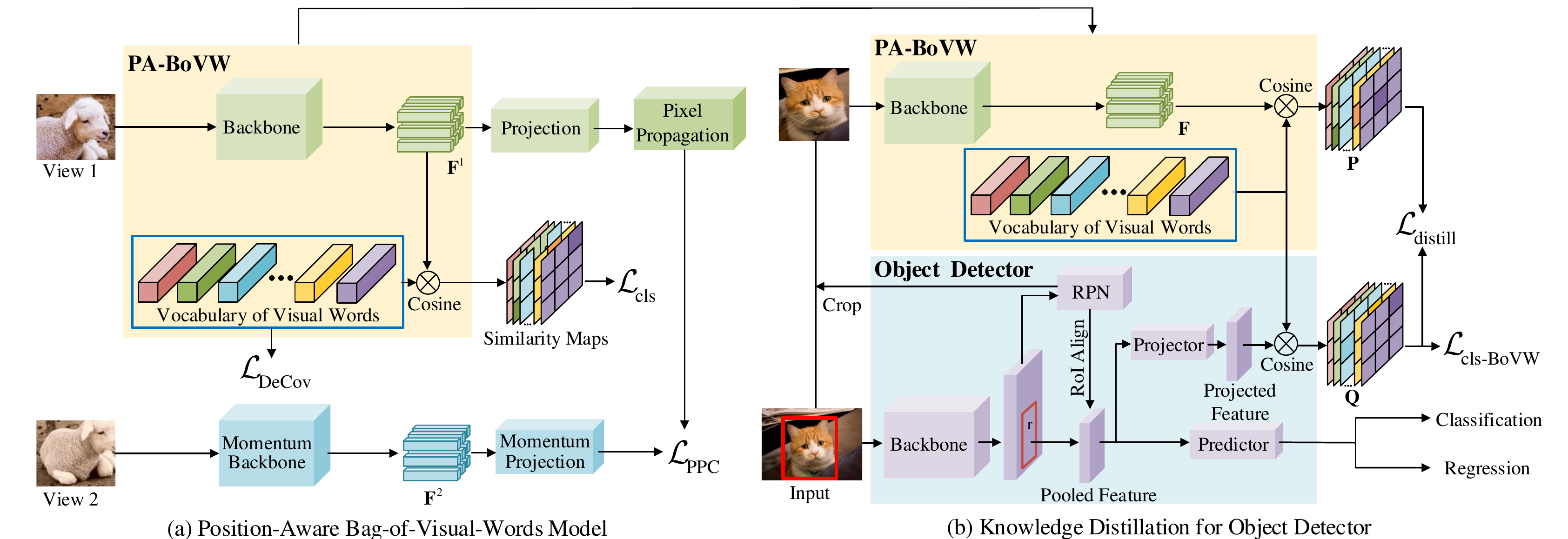}
   \caption{The overall architecture of our method. We first train the proposed \emph{PA-BoVW} model for learning a bag of visual words (\emph{BoVW}) via two pretext tasks: pixel-to-propagation consistency~\cite{xie2021propagate} and base-class recognition task. During the training procedure of the detector, given a positive region proposal, we crop it from original images and fed it into our pre-learned \emph{PA-BoVW} model to obtain the \emph{BoVW} representation. Then we use the obtained \emph{BoVW} representation as distilled knowledge to guide the learning of the detector.}
   \label{fig:framework}
\end{figure}

To deal with the potential overfitting in few-shot object detection based on deep learning networks, we propose to perform knowledge distillation to guide the learning process of few-shot object detection. Specifically, we learn a bag of visual words (\emph{BoVW}) for encoding images. The learned visual words are presumably representative, hence an image should have consistent \emph{BoVW} representations in two independent feature spaces. We first pre-learn a feature space and the derived \emph{BoVW} representation in this space can be considered as distilled knowledge that is transferred to the learning of the few-shot object detector. The extracted features by the object detector for each positive proposal are expected to derive consistent \emph{BoVW} representation with the distilled knowledge, thereby avoiding overfitting during supervised learning.

In this section, we will first elaborate on the proposed Position-Aware Bag-of-Visual-Words model (\emph{PA-BoVW}) for learning a bag of visual words and encoding images based on these visual words. Then we will show how to perform knowledge distillation to guide the learning of few-shot object detection.

\subsection{Position-Aware Bag-of-Visual-Words Model}
A typical way of constructing a bag of visual words from an image corpus is to cluster the image patches in the corpus in deep feature space and select the clustering centroids as the visual words~\cite{gidaris2020learning,jain2020quest}. While such an unsupervised method is straightforward and feasible given a sufficiently large image corpus, it shows limited effectiveness when the size of image corpus is limited. This is mainly because such a method tends to focus on the statistically frequent image patches which may not be semantically representative.

In this work we present a novel Position-Aware Bag-of-visual-Words (\emph{PA-BoVW}) model, which is able to learn a representative vocabulary of visual words from a limited size of image set. As shown in Figure~\ref{fig:framework} (a), we view each visual word as a learnable embedding and learn the parameters of all word embeddings in a supervised way based on two pretext tasks. The first pretext task is Pixel-to-Propagation Consistency~\cite{xie2021propagate}, which trains the backbone network to construct the embedding space for learning visual words in a self-supervised learning framework. Then we perform image classification on the representations encoded by our \emph{PA-BoVW} model. Hence, the task of image classification serves as the second pretext task for optimizing the parameter learning of both the word embeddings and the backbone network. To this end, we first construct an iconic-object image dataset $\mathcal{D}$ as image corpus by extracting all base class objects from the detection dataset according to their ground-truth bounding boxes and labels. We then train our \emph{PA-BoVW} model on $\mathcal{D}$.

\noindent\textbf{Learnable word embeddings.} Unlike the typical way that identifies visual words as the clustering centroids in the deep feature space, we learn visual words as learnable vectorial embeddings. A prominent benefit of such way is that the visual words do not necessarily correspond to image patches in the corpus. Instead, our model can learn an effective vocabulary of word embeddings freely from the whole embedding space under the optimization of the designed supervision (i.e., the classification pretext-task in our case).

\noindent\textbf{Self-supervised learning of embedding space via Pixel-to-Propagation Consistency.} 
To construct an effective embedding space for learning visual words, we employ a backbone network to encode the input image into a latent feature space and optimize the backbone network using Pixel-to-Propagation Consistency (PPC)~\cite{xie2021propagate} in a self-supervised way. PPC optimizes the backbone to make each pixel distinguishable from other pixels in the embedding space. 

Formally, given an image corpus $\mathcal{D}$, for an image $I \in \mathcal{D}$, two views ($I^1, I^2$) are generated by typical data augmentations (random cropping, color distortion, etc.). They are then fed into a self-supervised framework including a regular encoding network and a paired momentum encoding network to extract features respectively, as shown in Figure~\ref{fig:framework} (a). The encoding network consists of a backbone network and a projection network. The features which pass the backbone networks are denoted as ($\mathbf{F}^1, \mathbf{F}^2$). Then the projection networks convert them to ($\mathbf{E}^1, \mathbf{E}^2$). Each pixel in the feature maps that pass the regular encoding network are processed by a pixel propagation module~\cite{xie2021propagate} to enrich its feature by attending to all other pixels in the same view according to their similarities, for instance, the pixel $\mathbf{x}_i^1$ in $\mathbf{E}^1$ is processed as:
\begin{equation}
    \small
  q(\mathbf{x}_{i}^1) = \sum \nolimits_{j\in \mathbf{E}^1} \max(\frac{  (\mathbf{x}_i^1)^\top \cdot \mathbf{x}_j^1  }{\left\| \mathbf{x}_i^1 \right\|_2 \left\| \mathbf{x}_j^1 \right\|_2  }, 0)^2 \cdot \Psi(\mathbf{x}_j^1), \forall \mathbf{x}_{i}^1 \in \mathbf{E}^1.
  \label{eq:pixpro}
\end{equation}
$\Psi(\cdot)$ is a feature transformation module comprising 2 convolution layers with a batch normalization layer and a ReLU layer. The obtained feature $q(\mathbf{x}_{i}^1)$ is then used to maximize its cosine similarities with its corresponding pixel $\mathbf{x}_i^2$ in the other view $\mathbf{E}^2$ passing the momentum backbone:
\begin{equation}
  \mathcal{L}_{\text{PPC}} = 2 - \frac{  q(\mathbf{x}_i^1)^\top \cdot \mathbf{x}_{i}^2  }{\left\| q(\mathbf{x}_i^1) \right\|_2 \left\| \mathbf{x}_{i}^2 \right\|_2 } - \frac{q(\mathbf{x}_i^2)^\top \cdot \mathbf{x}_{i}^1  }{\left\| q(\mathbf{x}_i^2) \right\|_2 \left\| \mathbf{x}_{i}^1 \right\|_2 }.
  \label{eq:l_pixpro}
\end{equation}
Note that $\mathbf{x}_i^1$ and $\mathbf{x}_i^2$ are the corresponding pixels in two views ($\mathbf{E}^1, \mathbf{E}^2$), but they could have different positional coordinates. Since only the feature maps going through the regular backbone are processed by PPC~\cite{xie2021propagate}, both views have chance to go through the regular backbone and the momentum backbone respectively.

Since the embedding space is constructed independently from the learning of the object detector, the embedding space and the feature space of the object detector are two different feature spaces. Thus, the constructed embedding space can be used to not only learn the vocabulary of visual words, but encode \emph{BoVW} representations based on the learned visual words for an image as distilled knowledge. Such distilled knowledge is further used to guide the learning of the object detector.

\noindent\textbf{Position-aware encoding of \emph{BoVW} representation.}
Typical way of representing images using the visual words is to divide an image into patches and calculate the histogram over the visual words~\cite{lazebnik2006beyond}. However, such encoding scheme loses the position information which is crucial for object detection. To address this limitation, we encode an image using the vocabulary of visual words by calculating the Cosine similarity between each pixel of the image and each visual word in the embedding space while retaining the positional relationship among pixels. Formally, given an image $I$, its features which pass the backbone network are denoted as $\mathbf{F}\in \mathbb{R}^{C\times H \times W}$ with $C$ feature maps of size $H \times W$. The learned vocabulary of visual words are denoted as $\mathbf{V}\in \mathbb{R}^{K \times D}$ in which $K$ is the number of visual words and $D$ is the feature dimension for each word. The Cosine similarity between the pixel $(h, w)$ in $\mathbf{F}$ and the $j$-th word in $\mathbf{V}$ is calculated as:
\begin{equation}
\mathbf{P}_{j,h,w}=\frac{\mathbf{V}_{j}^\top \cdot \mathcal{F}_{\text{conv}}(\mathbf{F}_{h,w})}{\left\|\mathbf{V}_{j}^\top \right\|_2 \left\| \mathcal{F}_{\text{conv}}(\mathbf{F}_{h,w}) \right\|_2  },
  \label{eq:encoding_bovw}    
\end{equation}
where $\mathcal{F}_{\text{conv}}$ is a transformation function implemented by a convolutional layer to project $\mathbf{F}$ from $C$ channels to $D$ channels. Consequently, we obtain a similarity map $\mathbf{P}\in \mathbb{R}^{K\times H \times W}$ as the encoded \emph{BoVW} representation for the image $I$, which retains the position information for each pixel.

\noindent\textbf{Supervised learning of visual words by the pretext task of image classification.}
We employ image classification as a pretext task to guide the learning of the visual words based on two considerations: 1) the visual words should be discriminative for object recognition in that our goal is object detection; 2) the learned visual words should have well generalizability and can be used for knowledge distillation to restrain potential overfitting for object detection, thus the pretext task should be independent of the task of object detection.

Given an image $I\in \mathcal{D}$, we first encode it with \emph{BoVW}. Then the obtained \emph{BoVW} representation is fed into a simple classification head consisting of a pooling layer and a fully-connected layer. Formally, we perform average pooling over the \emph{BoVW} representation of $I$, namely the similarity map $\mathbf{P}$, along the $H$ and $W$ dimension and thus achieve a vectorial representation whose dimension is equal to the size of the vocabulary $\mathbf{V}$:
\begin{equation}
  \mathbf{P}_\text{avg} = \frac{1}{HW}\sum_{h=1}^{H}\sum_{w=1}^{W}\mathbf{P}_{j,h,w}.
  \label{eq:avg_pool}
\end{equation}
The obtained $\mathbf{P}_\text{avg}$ are further fed into a fully-connected layer $\mathcal{F}_{\text{fc}}$ and Softmax function $\mathcal{F}_{\text{softmax}}$ for classification prediction. Cross Entropy (CE) loss is used for optimization:
\begin{equation}
    \mathcal{L}_\text{cls} = \text{CE}(y_I, \mathcal{F}_{\text{softmax}}(\mathcal{F}_{\text{fc}}(\mathbf{P}_\text{avg}))),
\end{equation}
where $y_I$ is the groundtruth label for image $I$.

\noindent\textbf{Reducing inter-word correlation.} 
To encourage the diversity of visual words and reduce the redundancy among words, we add DeCov loss~\cite{cogswell2015reducing} as an auxiliary loss to minimize the correlation between different visual words:
\begin{equation}
  \mathcal{L}_\text{DeCov} = \frac{1}{2}(\left\|  \Sigma(\mathbf{V}) \right\|^2_F - \left\| \text{diag}(\Sigma(\mathbf{V})) \right\|^2_2), \\
  \label{eq:sigmaV}
\end{equation}
where $\Sigma(\cdot)$ denotes the covariance matrix and $\text{diag}(\cdot)$ extracts the diagonal elements of a matrix.

The whole Bag-of-Visual-Words model can be trained under the supervision by above three loss functions jointly:
\begin{equation}
  \mathcal{L}_\text{BoVW} = \mathcal{L}_\text{PPC} + \mathcal{L}_\text{cls} + \mathcal{L}_\text{DeCov},
  \label{eq:l_total}
\end{equation}
Note that although our \emph{PA-BoVW} model is learned independently from the object detector in an extra step, it can be trained quite efficiently due to small size of object images and relatively simple supervision tasks compared to the task of object detection.

\subsection{Knowledge Distillation for Object Detection}
Our Position-Aware Bag-of-Visual-Words (\emph{PA-BoVW}) model learns the embedding space for visual words using PPC~\cite{xie2021propagate} as the pretext task, and learns the vocabulary of visual words based on the pretext task of image classification. Thus, our \emph{PA-BoVW} model is optimized in a completely independent way from the task of object detection. As a result, the encoded \emph{BoVW} representation in the embedding space of the \emph{PA-BoVW} model for an image can be viewed as distilled knowledge, which can be transferred to the learning process of few-shot object detection to suppress potential overfitting on this image. The rationale behind this design is that a well learned (non-overfitting) feature representation for an object by a detector should bear consistent similarity distribution over the learned visual words with the corresponding \emph{BoVW} representation by our \emph{PA-BoVW} model. Thus, we can derive consistent \emph{BoVW} representations from the learned features by the object detector and our \emph{PA-BoVW} model, respectively.

As shown in Figure~\ref{fig:framework} (b), we adopt the typical object detection framework, which is built upon Faster R-CNN~\cite{ren2015faster}. Actually, our proposed method of knowledge distillation can be readily integrated into any classical object detection framework. Given a positive region proposal $r$ generated by RPN (Region Proposal Network), which is assigned with one of the ground-truth labels and bounding boxes during training, we crop the corresponding region from the original input image and resize it to a fixed size by bilinear interpolation, then fed it into our \emph{PA-BoVW} model to obtain its \emph{BoVW} representation $\mathbf{P}(r)\in \mathbb{R}^{K\times H \times W}$ by Eq.~\ref{eq:encoding_bovw}. Meanwhile, we calculate the Cosine similarities between the features $\mathbf{G}(r)$ for $r$, obtained from the RoI pooling layer of the object detector, and the vocabulary of visual words $\mathbf{V}$. Note that $\mathbf{G}(r)$ and the visual words are not learned in the same feature space, thus we project them into the same feature space first and then compute the Cosine similarities in the same way as Eq.\ref{eq:encoding_bovw}:
\begin{equation}
\begin{split}
  & \mathbf{Q}_{j,h,w}(r) = \frac{g(\mathbf{V}_j)^\top \cdot \phi(\mathbf{G}_{h,w}(r))}{\left\| g(\mathbf{V}_j) \right\|_2  \left\| \phi(\mathbf{G}_{h,w}(r)) \right\|_2}, \\
  & j=1, \dots, K, h=1,\dots,H, w=1, \dots, W.
  \label{eq:student}
  \end{split}
\end{equation}
Herein, $g(\cdot)$ is the project function implemented as a fully connected layer for $\mathbf{V}$, while $\phi(\cdot)$ denotes the project function for features $\mathbf{G}(r)$, which is formulated as a $1\times1$ convolutional layer. $K, H, W$ are the size of the vocabulary $\mathbf{V}$, the height and the width of $\mathbf{G}(r)$, respectively.

The obtained $\mathbf{Q}(r)\in \mathbb{R}^{K\times H \times W}$ is equivalent to the \emph{BoVW} representation encoded on the learned features $\mathbf{G}$ of the object detector. If $\mathbf{G}(r)$ is learned well and not overfitting on the input data, it should result in consistent \emph{BoVW} representation as our pre-trained \emph{PA-BoVW} model. Thus, we minimize the L1-norm distance between these two \emph{BoVW} representations to guide the learning process of the object detector:
\begin{equation}
  L_\text{distill} = \frac{1}{RHW}\sum^{R}_{r=1}\sum^{H}_{h=1}\sum^{W}_{w=1}\left\| \mathbf{P}_{h,w}(r)-\mathbf{Q}_{h,w}(r) \right\|_1,
  \label{eq:distillation}
\end{equation}
where $R$ is the number of positive proposals.

\noindent\textbf{Knowledge distillation on both base classes and novel classes.} 
As most fine-tuning based methods~\cite{qiao2021defrcn,sun2021fsce,wang2020frustratingly,wu2020multi} do, we first train the object detector on base classes and then fine-tune the model on the novel classes. We perform the knowledge distillation process in both training stages to restrain the potential overfitting on base classes and novel classes, respectively.

\noindent\textbf{Collaborative object detection with \emph{BoVW} representations.} 
The obtained \emph{BoVW} representation $\mathbf{Q}(r)$ can also be used for object classification for the region proposal $r$. Thus, we perform classification by fusing the predicted scores from $\mathbf{Q}(r)$ and the original features respectively:
\begin{equation}
\begin{split}
  & p = \eta \cdot p_\text{orig} + (1 - \eta) \cdot p^{\prime},\\
  & p^{\prime} = \mathcal{F}_{\text{softmax}}(\mathcal{F}_{\text{fc}}(\mathbf{Q}(r))),
  \end{split}
  \label{eq:score}
\end{equation}
where $p_\text{orig}$ and $p^{\prime}$ are predicted scores from the original features and from the \emph{BoVW} representation respectively. Here $p^{\prime}$ is obtained by performing a linear transformation $\mathcal{F}_\text{fc}$ and softmax function on $\mathbf{Q}(r)$. $\eta$ is a hyper-parameter to fuse two scores. During training, Cross Entropy loss is used as an auxiliary loss to guide the optimization:
\begin{equation}
    \mathcal{L}_\text{cls-BoVW} = \text{CE}(y, p^{\prime}),
\end{equation}
where $y$ is the groundtruth label for the region proposal $r$.

Consequently, the object detector is trained under the supervision of three losses jointly:
\begin{equation}
  \mathcal{L}_\text{obj} = \mathcal{L}_\text{det} + \mathcal{L}_\text{distill} + \mathcal{L}_\text{cls-BoVW},
  \label{eq:loss_student}
\end{equation}
where $\mathcal{L}_{det}$ corresponds to the standard Faster R-CNN~\cite{ren2015faster} losses for object detection, including the losses for RPN, classification and box regression.

\section{Experiments}
\label{sec:exp}
\subsection{Experimental Setup}
\label{setting}
\noindent\textbf{Benchmarks.}
We evaluate our approach on PASCAL VOC~\cite{everingham2010pascal} dataset and MS COCO~\cite{lin2014microsoft} dataset. We follow the previous work~\cite{kang2019few} for data construction to have a fair comparison. PASCAL VOC comprises 15 base classes and 5 novel classes. We utilize the same three class splits introduced in~\cite{kang2019few}, where each novel class has $k=1, 2, 3, 5, 10$ instances sampled from the combination of VOC 2007 and VOC 2012 trainval sets. VOC 2007 test set is used for evaluation. As for MS COCO, the 60 categories disjoint with PASCAL VOC are selected as base classes, and the remaining 20 categories are used as novel classes with $K=10, 30$. For evaluation metrics, we report AP50 of novel classes (nAP50) for PASCAL VOC and COCO-style AP of the novel classes for MS COCO.

\noindent\textbf{Implementation Details.}
We evaluate our approach by building it upon two state-of-the-art methods: TFA++~\cite{sun2021fsce} and DeFRCN~\cite{qiao2021defrcn}. TFA++~\cite{sun2021fsce} is a strong baseline which jointly fine-tunes the feature extractors and box predictors during the fine-tuning stage. DeFRCN~\cite{qiao2021defrcn} is a simple yet effective architecture which is the current state of the art.

For $\emph{PA-BoVW}$ model, we use an ImageNet~\cite{russakovsky2015imagenet} pre-trained ResNet101~\cite{he2016deep} as the backbone. The input size is $224 \times 224$. The feature dimension of visual word is 512. The number of visual words is set to 256 for PASCAL VOC and 1024 for MS COCO.
We follow the same data augmentation strategy in PPC~\cite{xie2021propagate}, where two random patches of an image are independently sampled, followed by random horizontal flip, color distortion, gaussian blur, and solarization. We use AdamW optimizer to optimize the $\emph{PA-BoVW}$ model with the initial learning rate of 1e-4 for 24 epochs. We decay the learning rate by ratio 0.1 at epoch 18 and 22. The total batch size is set to 256. The object detector is trained on 8 GPUs with a batch size of 16. The $\eta$ is uniformly set to 0.5. All other training settings are the same as that in TFA++\cite{sun2021fsce} and DeFRCN\cite{qiao2021defrcn}.

\setlength\tabcolsep{0.9pt}
\begin{table}[!tbp]
  \centering
  \caption{Comparison with existing few-shot object detection methods using nAP50 as evaluation metric on three PASCAL VOC Novel Split sets. `Ours (KD-TFA++)' denotes our method using TFA++~\cite{sun2021fsce} as the baseline. $\dagger$ indicates that model is evaluated using the released code.}
    \scriptsize
  \begin{tabular}{l| c c c c c | c c c c c | c c c c c}
    \toprule
    \multirow{2}{*}{Method / Shots}  & \multicolumn{5}{c|}{Novel Split 1} & \multicolumn{5}{c|}{Novel Split 2} & \multicolumn{5}{c}{Novel Split 3} \\
    & 1 & 2 & 3 & 5 & 10 & 1 & 2 & 3 & 5 & 10& 1 & 2 & 3 & 5 & 10 \\
    \midrule
    LSTD~\cite{chen2018lstd}      &8.2&1.0&12.4&29.1&38.5&11.4&3.8&5.0&15.7&31.0&12.6&8.5&15.0&27.3&36.3\\
    MetaDet~\cite{wang2019meta} &18.9&20.6&30.2&36.8&49.6&21.8&23.1&27.8&31.7&43.0&20.6&23.9&29.4&43.9&44.1   \\
    Meta R-CNN~\cite{yan2019meta}  &19.9&25.5&35.0&45.7&51.5&10.4&19.4&29.6&34.8&45.4&14.3&18.2&27.5&41.2&48.1 \\
    FSRW~\cite{kang2019few}   &14.8&15.5&26.7&33.9&47.2&15.7&15.3&22.7&30.1&40.5&21.3&25.6&28.4&42.8&45.9   \\
    RepMet~\cite{karlinsky2019repmet}      &26.1&32.9&34.4&38.6&41.3&17.2&22.1&23.4&28.3&35.8&27.5&31.1&31.5&34.4&37.2\\
    NP-RepMet~\cite{yang2020restoring}      &37.8&40.3&41.7&47.3&49.4&\textbf{41.6}&43.0&43.4&47.4&49.1&33.3&38.0&39.8&41.5&44.8\\
    MPSR~\cite{wu2020multi}      &41.7&-&51.4&55.2&61.8&24.4&-&39.2&39.9&47.8&35.6&-&42.3&48.0&49.7\\
    TFA w/cos~\cite{wang2020frustratingly}         &39.8&36.1&44.7&55.7&56.0&23.5&26.9&34.1&35.1&39.1&30.8&34.8&42.8&49.5&49.8 \\
    HallucFsDet~\cite{zhang2021hallucination}  &47.0&44.9&46.5&54.7&54.7&26.3&31.8&37.4&37.4&41.2&40.4&42.1&43.3&51.4&49.6\\
    Retentive R-CNN\cite{fan2021generalized} &42.4&45.8&45.9&53.7&56.1&21.7&27.8&35.2&37.0&40.3&30.2&37.6&43.0&49.7&50.1\\
    FSCE~\cite{sun2021fsce}  &44.2&43.8&51.4&61.9&63.4&27.3&29.5&43.5&44.2&50.2&37.2&41.9&47.5&54.6&58.5\\
    FADI~\cite{fadi2021}&50.3&54.8&54.2&59.3&63.2&30.6&35.0&40.3&42.8&48.0&45.7&49.7&49.1&55.0&59.6\\
    CME~\cite{li2021beyond} &41.5&47.5&50.4&58.2&60.9&27.2&30.2&41.4&42.5&46.8&34.3&39.6&45.1&48.3&51.5\\
    UP-FSOD~\cite{wu2021universal} &43.8&47.8&50.3&55.4&61.7&31.2&30.5&41.2&42.2&48.3&35.5&39.7&43.9&50.6&53.3\\
    QA-FewDet~\cite{han2021query}&42.4&51.9&55.7&62.6&63.4&25.9&37.8&46.6&48.9&51.1&35.2&42.9&47.8&54.8&53.5\\
    \midrule
    TFA++$^{\dagger}$~\cite{sun2021fsce}&43.4&42.1&47.3&57.2&60.8&24.3&27.7&42.0&42.0&48.5&38.0&41.0&45.8&54.0&56.2\\
    \rowcolor{mygray} Ours (KD-TFA++)&47.0&50.2&52.5&62.1&64.2&29.7&32.9&45.9&48.5&51.1&42.6&46.5&48.8&56.8&57.4 \\
    \midrule
     DeFRCN~\cite{qiao2021defrcn}&57.0&58.6&64.3&67.8&67.0&35.8&42.7&51.0&54.5&52.9&52.5&56.6&55.8&60.7&\textbf{62.5} \\
    \rowcolor{mygray} Ours (KD-DeFRCN)& \textbf{58.2}&\textbf{62.5}&\textbf{65.1}&\textbf{68.2}&\textbf{67.4}&37.6&\textbf{45.6}&\textbf{52.0}&\textbf{54.6}&\textbf{53.2}&\textbf{53.8}&\textbf{57.7}&\textbf{58.0}&\textbf{62.4}&62.2 \\
    \bottomrule
  \end{tabular}
  \label{tab:voc}
\end{table}
\subsection{Comparison with State-of-the-art Methods}
\label{comparison}
\noindent\textbf{Results on PASCAL VOC.}
Table~\ref{tab:voc} presents the results on PASCAL VOC, which show that our approach improves the performance of TFA++\cite{sun2021fsce} by a large margin in all cases including different numbers of training shots in different splits. Particularly, for the 2-shot case of Novel Split 1, 5-shot case of Novel Split 2 and 2-shot case of Novel Split 3, our approach is 8.1\%, 6.5\%, 5.5\% higher than the baseline. When applying our approach to DeFRCN\cite{qiao2021defrcn}, which is the current state of the art, our method still improves the performance in most cases, especially in the extremely-few-shot regimes such as 1-shot and 2-shot.

\noindent\textbf{Results on MS COCO.}
Table~\ref{tab:coco} shows the results on MS COCO. Applying our approach to two baselines achieves 1.1\% and 0.3\% nAP performance gain for 10-shot, 0.6\% and 0.1\% in terms of novel AP performance gain for 30-shot, respectively. There is no as large performance gain as on PASCAL VOC, which is probably because MS COCO has much more training images and thus has a lower risk of overfitting. To validate this speculation, we further evaluate our method by only using a small subset of base-class data for training. Specifically, we randomly select 10\% samples from base classes to form a training set. For novel classes, we keep the same setting in standard few-shot object detection. Table~\ref{tab:coco_10} shows that the performance gains are larger than using all training data.

\begin{table}[t]
\begin{minipage}[!t]{0.46\linewidth}
\setlength\tabcolsep{1.2pt}
\centering
  \caption{Few-shot object detection performance on MS COCO.}
    \scriptsize
  \begin{tabular}{l| c c | c c}
    \toprule
    \multirow{2}{*}{Method}  & \multicolumn{2}{c|}{nAP} & \multicolumn{2}{c}{nAP75} \\
    & 10 & 30 & 10 & 30 \\
    \midrule
    LSTD~\cite{chen2018lstd}     &3.2&6.7&2.1&5.1\\
    MetaDet~\cite{wang2019meta}  &7.1&11.3&6.1&8.1\\
    Meta R-CNN~\cite{yan2019meta}&8.7&12.4&6.6&10.8\\
    FSRW~\cite{kang2019few}     &5.6&9.1&4.6&7.6\\
    TFA w/cos~\cite{wang2020frustratingly} &10.0&13.7&9.3&13.4\\
    MPSR~\cite{wu2020multi}      &9.8&14.1&9.7&14.2\\
    SRR-FSD~\cite{zhu2021semantic}   &11.3&14.7&9.8&13.5\\
    Retentive R-CNN~\cite{fan2021generalized}&10.5&13.8&$-$&$-$\\
    FSCE~\cite{sun2021fsce}      &11.9&16.4&10.5&16.2  \\
    FADI~\cite{fadi2021}&12.2&16.1&11.9&15.8\\
    CME~\cite{li2021beyond}       &15.1&16.9&16.4&17.8\\
    UP-FSOD~\cite{wu2021universal}&11.0&15.6&10.7&15.7\\
    QA-FewDet~\cite{han2021query}&11.6&16.5&9.8&15.5\\
    \midrule
    TFA++$^{\dagger}$~\cite{sun2021fsce}&11.7&16.0&10.3&15.3  \\
    \rowcolor{mygray} Ours (KD-TFA++)&12.8&16.6&11.5&16.1\\
    \midrule
    DeFRCN~\cite{qiao2021defrcn}&18.6&22.5&17.6&22.3  \\
    \rowcolor{mygray}
    Ours (KD-DeFRCN)&\textbf{18.9}&\textbf{22.6}&\textbf{17.8}&\textbf{22.6}\\
    \bottomrule
  \end{tabular}
  \label{tab:coco}
\end{minipage}
\begin{minipage}[!t]{0.52\linewidth}
    \begin{minipage}[!t]{1.0\linewidth}
    \setlength\tabcolsep{2.2pt}
  \centering
  \caption{Results on MS COCO with 10\% labeled base-class samples.}
    \scriptsize
  \begin{tabular}{l| c c | c c}
    \toprule
    \multirow{2}{*}{Method}  & \multicolumn{2}{c|}{nAP} & \multicolumn{2}{c}{nAP75} \\
    & 10 & 30 & 10 & 30 \\
    \midrule
    DeFRCN~\cite{qiao2021defrcn}&12.1&14.9&8.5&11.5  \\
    Ours (KD-DeFRCN)&\textbf{13.0}&\textbf{16.0}&\textbf{9.7}&\textbf{12.6}\\
    \bottomrule
  \end{tabular}
  \label{tab:coco_10}
    \end{minipage}
    ~\\
    ~\\
    ~\\
    \begin{minipage}[t]{1.0\linewidth}
    \setlength\tabcolsep{1.0pt}
\centering
  \caption{Effect of each loss function.}
    \scriptsize
  \begin{tabular}{ c | c c | c c c }
    \toprule
    $L_\text{DeCov}$ & $L_\text{distill}$ & $L_\text{cls-BoVW}$ & 3-shot & 5-shot & 10-shot \\
    \midrule
    $\checkmark$ &&&47.3&57.2&60.8\\
    $\checkmark$ & $\checkmark$ & & 51.2 & 59.9 & 62.6 \\
    & $\checkmark$ & $\checkmark$ & 50.8 & 60.1 & 61.3 \\
    $\checkmark$ & $\checkmark$ & $\checkmark$ & \textbf{51.6} & \textbf{60.6} & \textbf{63.1} \\
    \bottomrule
  \end{tabular}
  \label{tab:loss_function}
    \end{minipage}
\end{minipage}
\end{table}

\subsection{Ablation Studies}
\label{ablation}
In this section, we conduct ablation studies on the Novel Split 1 of PASCAL VOC using TFA++~\cite{sun2021fsce} as the baseline. 

\noindent\textbf{Effect of each functional component.} Table~\ref{tab:components} shows the efficacy of each functional components for few-shot object detection on novel classes, including distillation on base classes, novel classes and score fusion for classification in Equation~\ref{eq:score}.  With the knowledge distillation for base classes, the performance gain is 4.2\%/2.5\%/1.9\% for 3/5/10-shot, respectively. By performing distillation on novel class during the fine-tuning stage, the performance gain increases 0.1\%/0.9\%/0.4\% respectively, which indicates the improved generalization ability on novel classes
by our method. Finally, fusing the predicted scores from the original features and from the BoVW representation for classification yields the extra performance gain by 0.9\%/1.5\%/1.1\%.

\noindent\textbf{Effect of each loss function.}
Table~\ref{tab:loss_function} shows the effect of each loss function. Both the distillation loss $L_\text{distill}$ and collaborative detection loss $\mathcal{L}_\text{cls-BoVW}$ improve the performance distinctly. Comaring the results in the last two rows, the $\mathcal{L}_\text{DeCov}$ which is designed to encourage the diversity of visual words and reduce the redundancy among words, also improves the performance.

\noindent\textbf{Quantification for the overfitting on base classes.} 
Fine-tuning training strategy tends to make models overfit on base classes. Since most parameters of the feature extractors are freezed or just fine-tuned slightly during the fine-tuning stage, most model capacity is allocated to fitting the base samples. To quantify such overfitting on base classes, we perform three experiments:
1) Using fine-tuning training strategy, we first pre-train the baseline TFA++ on base classes, then fine-tune it with sufficient novel-class samples instead of k shot per class, which is denoted as two-stage training mode; 2) Similar to the setting in 1), we fine-tune our model with sufficient novel-class samples after pre-training; 3) we train the baseline using sufficient samples for both base and novel classes together in one-stage mode. We compare both nAP50 and the number of misclassified samples from novel to base classes to measure the overfitting. The results in Table~\ref{tab:overfitting} show that 1) the baseline trained in two-stage mode performs much worse than training itself in one-stage mode and has a larger number of misclassified samples, indicating that the model is heavily biased towards base classes; 2) using the same two-stage training mode, our method achieves 3.5\% of performance gain than baseline and substantially decreases misclassified cases, which demonstrates the effectiveness of our method for suppressing the overfitting on base classes.

\setlength\tabcolsep{5pt}
\begin{table}[!t]
  \centering
  \caption{Ablation studies of key components.}
        \renewcommand\arraystretch{1.0}
    \scriptsize
  \begin{tabular}{c c c c| c c c}
    \toprule
    \multirow{2}{*}{Baseline} & \multirow{2}*{\shortstack{Distillation\\for Base}} & \multirow{2}*{\shortstack{Distillation\\for Novel}}& \multirow{2}*{\shortstack{Score\\Fusion}} &  \multicolumn{3}{c}{nAP50} \\
    &&&   & 3-shot & 5-shot & 10-shot \\
    \midrule
    $\checkmark$ & & & & 47.3 & 57.2 & 60.8 \\
    $\checkmark$ & $\checkmark$ & & &51.5&59.7&62.7 \\
    $\checkmark$ & $\checkmark$ & $\checkmark$ & & 51.6& 60.6& 63.1 \\
    $\checkmark$ & $\checkmark$ & $\checkmark$ & $\checkmark$ & \textbf{52.5} & \textbf{62.1} & \textbf{64.2} \\
    \bottomrule
  \end{tabular}
  \label{tab:components}
\end{table}

\setlength\tabcolsep{3.5pt}
\begin{table}[t]
  \centering
  \caption{Quantification for the overfitting.} 
        \renewcommand\arraystretch{1.0}
    \scriptsize
  \begin{tabular}{l| c | c}
    \toprule
    Methods & nAP50 & Misclassified cases (novel$\rightarrow$base)  \\
    \midrule
    TFA++ (two-stage)& 73.2 & 1021 \\
    Ours (KD-TFA++) (two-stage) & 76.7 & 723 \\
    \midrule
    TFA++ (one-stage) & 85.8 & 631 \\
    \bottomrule
  \end{tabular}
  \label{tab:overfitting}
\end{table} 

\begin{table}[t!]
\begin{minipage}[t]{0.46\linewidth}
\centering
\caption{Effect of different pretext tasks.}
\scriptsize
\begin{tabular}{l | c c c }
    \toprule
    \multirow{2}{*}{Pretext tasks} &  \multicolumn{3}{c}{nAP50} \\
     & 3 & 5 & 10 \\
    \midrule
    w/o distillation& 47.3 & 57.2 & 60.8 \\
    Cluster& 49.6 & 57.4 & 61.1\\
    Cls &52.3& 60.8&63.6 \\
    Cls + PPC &\textbf{52.5} & \textbf{62.1} & \textbf{64.2} \\
    \bottomrule
  \end{tabular}
  \label{tab:bovw}
\end{minipage}
\setlength\tabcolsep{1.5pt}
\begin{minipage}[t]{0.50\linewidth}
\centering
\caption{Distillation on deep features vs on \emph{BoVW} representations.}
    \scriptsize
  \begin{tabular}{l| c c c c c }
    \toprule
    \multirow{2}{*}{\shortstack{Distillation\\methods}} &  \multicolumn{5}{c}{nAP50} \\
      & 1 & 2 & 3 & 5 & 10 \\
    \midrule
    Baseline & 43.4 & 42.1 & 47.3 & 57.2 & 60.8 \\
    On deep features & 33.2 & 37.9 & 36.5 & 47.1 & 48.3 \\
    On BoVW & \textbf{47.0} & \textbf{50.2} & \textbf{52.5} & \textbf{62.1} & \textbf{64.2} \\
    \bottomrule
  \end{tabular}
  \label{tab:method_distill}
\end{minipage}
\end{table}

\noindent\textbf{Effect of different pretext tasks for learning \emph{BoVW} models.}
We conduct experiments to compare our \emph{PA-BoVW} model and the typical method for learning the visual words (denoted as \emph{`Cluster'}), which trains a classification network on base classes and selects the clustering centroids of the feature vectors as visual words. Then we evaluate the performance our \emph{PA-BoVW} optimized using only the image classification as the pretext task (denoted as \emph{`Cls'}). 
Table~\ref{tab:bovw} shows that the performance gains from \emph{`Cluster'} is smaller than that of our \emph{PA-BoVW} using only the pretext task of image classification. Using the pretext task of PPC further boosts the performance substantially, which implies the importance of learning the embedding space by PPC in a self-supervised way. 

\noindent\textbf{Distillation on deep features vs on \emph{BoVW} representations}. A classical way to perform knowledge distillation is to learn an independent feature space and perform distillation between two feature spaces. We conduct such experiment to compare between distillation on deep features and on \emph{BoVW} representations. Specifically, we directly distill the pooled deep features in our trained \emph{PA-BoVW} model to the feature space of the detector. The results of such method shown in Table~\ref{tab:method_distill} are much worse than our method. This is reasonable since distillation between feature space relies heavily on 1) the quality of the referenced features from which the distillation is performed and 2) well modeling of mapping between two feature space. By contrast, our method distills knowledge based on the similarity distribution over learned visual words, which benefits from similar merits of feature representations as Bag-of-Word representations used in NLP.

\begin{figure}[t] \centering
\subfigure[t-SNE.] { \label{fig:tfa_tsne}
\includegraphics[width=0.42\columnwidth]{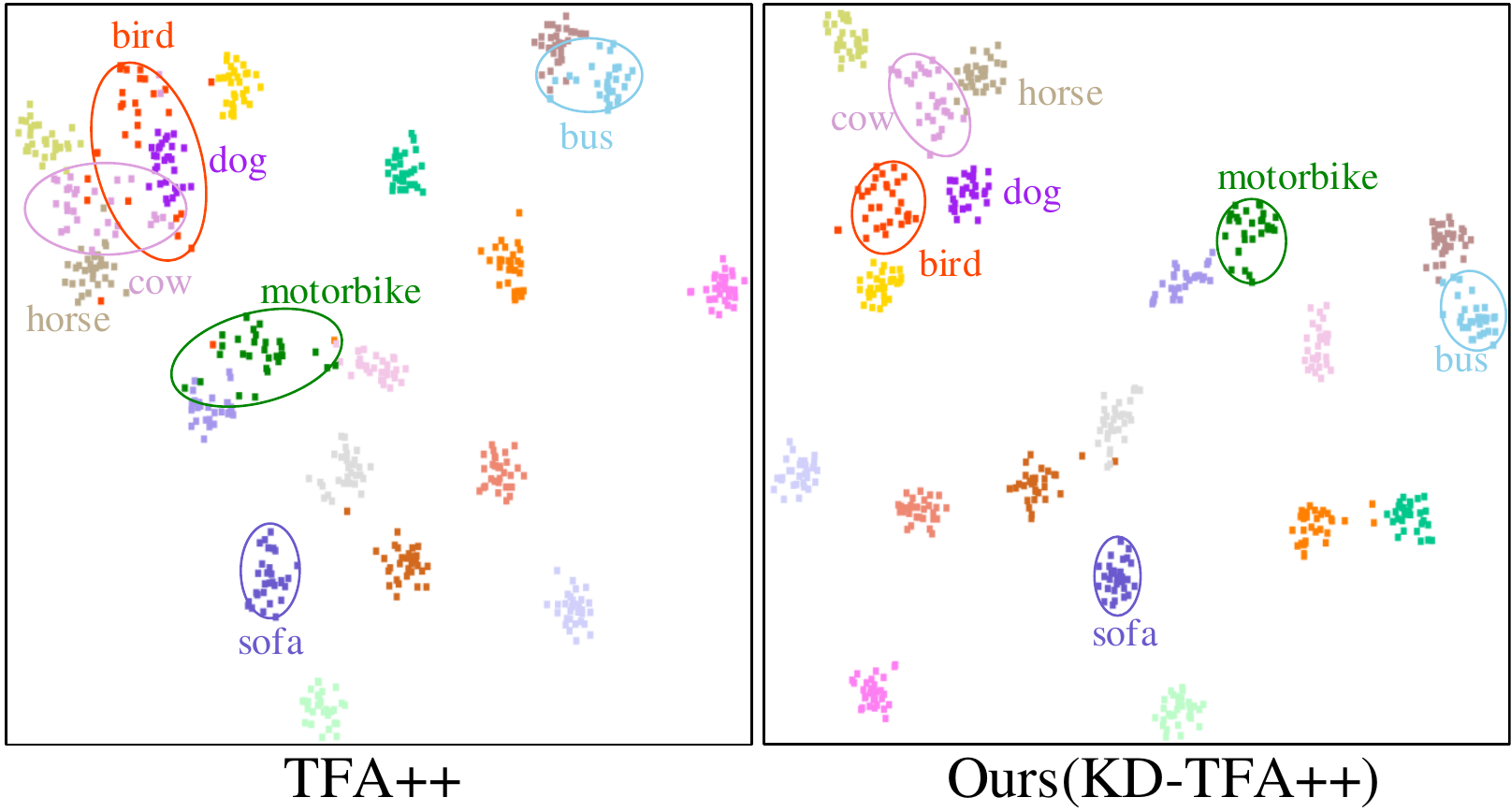}
}
\subfigure[Detection results.] { \label{fig:bow_tsne}
\includegraphics[width=0.5\columnwidth]{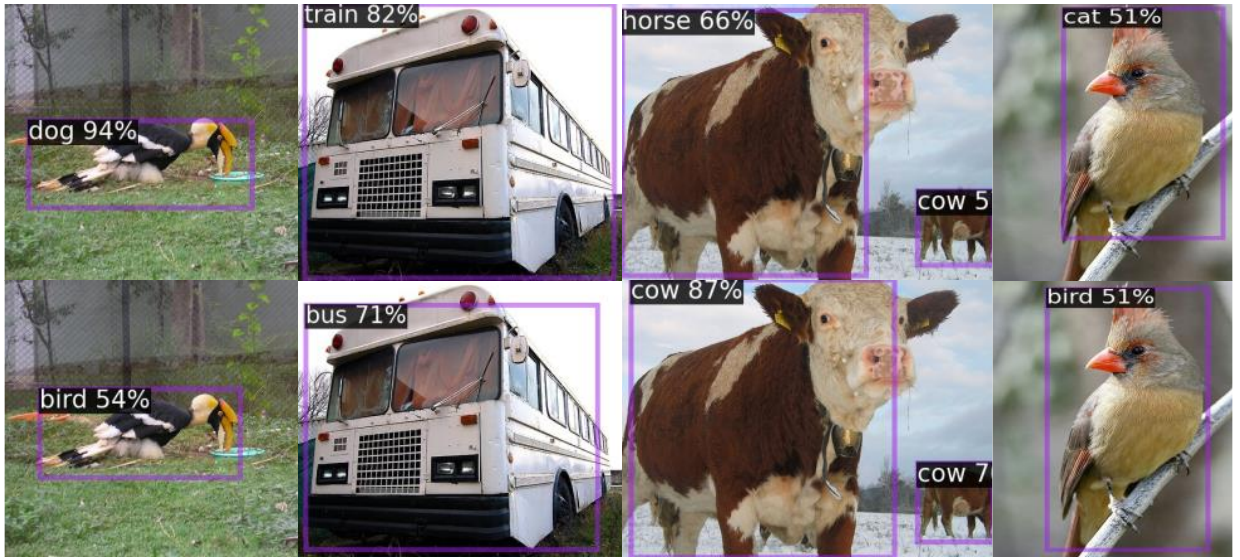}
}
\caption{(a) The t-SNE visualization of proposal embeddings of baseline with and without distillation. (b) Detection results based on the 10-shot case. The first row shows the results of the baseline and the second row shows the results of our approach.}
\label{fig:vis}
\end{figure}

\noindent\textbf{Qualitative Evaluation.}
Figure~\ref{fig:vis} (a) shows the t-SNE~\cite{van2008visualizing} visualization of proposal embeddings from randomly selected 30 instance bounding boxes per category. The baseline (TFA++~\cite{sun2021fsce}) tends to misclassify some samples of novel classes as similar base classes. For instance, the samples from novel classes `bird' and `cow' cannot be clearly separated from other base classes like `dog' and `horse'. In contrast, applying our approach to the baseline model leads to more accurate boundaries. Figure~\ref{fig:vis} (b) shows the detection results of the baseline and our approach. We can observe that our method can successfully detect the novel objects while the baseline tends to misclassify these objects as base classes.

\section{Conclusion}
To solve the potential overfitting in few-shot object detection, we propose a knowledge distillation framework. We first learn a $\emph{PA-BoVW}$ model using two pretext tasks, namely Pixel-to-Propagation Consistency and image classification. Based on the $\emph{PA-BoVW}$ model, then we perform distillation to guide the learning of detector. As an orthogonal component, our approach can be easily combined with other methods and significantly improve the performance.

\subsubsection{Acknowledgements} 
This work was supported in part by the NSFC fund (U2013210, 62006060, 62176077), in part by the Guangdong Basic and Applied Basic Research Foundation under Grant (2019Bl515120055, 2021A1515012528, 2022A1515010306), in part by the Shenzhen Key Technical Project under Grant 2020N046, in part by the Shenzhen Fundamental Research Fund under Grant (JCYJ20210324132210025), in part by the Shenzhen Stable Support Plan Fund for Universities (GXWD20201230155427003-20200824125730001, GXWD202012
30155427003-20200824164357001),  in part by the Medical Biometrics Perception and Analysis Engineering Laboratory, Shenzhen, China, and in part by the Guangdong Provincial Key Laboratory of Novel Security Intelligence Technologies (2022B1212010005).

\clearpage
%
%
\bibliographystyle{splncs04}
\bibliography{egbib}

\clearpage

\title{Supplementary Material for Few-Shot Object Detection by Knowledge Distillation Using Bag-of-Visual-Words Representations} 
\titlerunning{Knowledge Distillation Using Bag-of-Visual-Words Representations}
\author{}
\authorrunning{W. Pei and S. Wu et al.}
\institute{}

\maketitle

\appendix
\section{Results on Base Classes}
Table~\ref{tab:voc_base} shows the performance for base and novel classes on Novel Split 1 of PASCAL VOC dataset. Although AP for base classes (bAP50) is not our primary concern, our method makes competitive results. It can be observed that our method improves not only the performance of novel classes, but also the performance of base classes. These results demonstrate that our method can maintain the performance on previous knowledge without forgetting.
\setlength\tabcolsep{6pt}
\begin{table}[h]
    \scriptsize
  \centering
  \caption{Few-shot object detection results for base and novel classes on Novel Split 1 of PASCAL VOC dataset. $\dagger$ indicates that model is evaluated using the released code.}
  \begin{tabular}{c c | c c }
    \toprule
    Shots  & Method & bAP50 & nAP50 \\
    \midrule
    \multirow{7}{*}{3}& FRCN+ft-full~\cite{yan2019meta}  & 63.6 & 32.8 \\
                      & Meta R-CNN~\cite{yan2019meta}    & 64.8 & 35.0 \\
                      & Baseline-FPN~\cite{wu2020multi}  & 66.2 & 41.1 \\
                      & MPSR~\cite{wu2020multi}          & 67.8 & 51.4 \\
                      & TFA w/cos~\cite{wang2020frustratingly}    & \textbf{79.1} & 44.7 \\
                      & FSCE~\cite{sun2021fsce}          & 74.1 & 51.4 \\
                      & TFA++$^{\dagger}$~\cite{sun2021fsce}         & 75.4 & 47.3 \\
                      & Ours (KD-TFA++) & 76.4 & \textbf{52.5} \\
    \midrule
    \multirow{5}{*}{5}& Baseline-FPN~\cite{wu2020multi}  & 67.9 & 49.6 \\
                      & MPSR~\cite{wu2020multi}          & 68.4 & 55.2 \\
                      & TFA w/cos~\cite{wang2020frustratingly}     & 77.0 & 55.6 \\
                      & FSCE~\cite{sun2021fsce}          & 76.6 & 61.9 \\
                      & TFA++$^{\dagger}$~\cite{sun2021fsce}         & 77.7 & 57.2 \\
                      & Ours (KD-TFA++) & \textbf{79.0} & \textbf{62.1} \\
    \midrule
    \multirow{6}{*}{10}& FRCN+ft-full~\cite{yan2019meta} & 61.3 & 45.6 \\
                       & Meta R-CNN~\cite{yan2019meta}   & 67.9 & 51.5 \\
                       & Baseline-FPN~\cite{wu2020multi} & 70.0 & 56.9 \\
                       & MPSR~\cite{wu2020multi}         & 71.8 & 61.8 \\
                       & TFA w/cos~\cite{wang2020frustratingly}    & 78.4 & 56.0 \\
                       & TFA++$^{\dagger}$~\cite{sun2021fsce}        & 77.5 & 60.8 \\
                       & Ours (KD-TFA++) & \textbf{78.6} & \textbf{64.2} \\
    \bottomrule
  \end{tabular}
  \label{tab:voc_base}
\end{table}

\section{Results over Multiple Runs}
To eliminate the effect of sample variance introduced by the random selection of few-shot training samples, we fine-tune our model over 10 random selections of few-shot training samples independently for each experimental settings (including different novel splits and shot numbers), and obtain the average results on PASCAL VOC dataset. As shown in Table~\ref{tab:voc_multiple}, our method improves the performance of  TFA++~\cite{sun2021fsce} under all settings.
\setlength\tabcolsep{1.0pt}
\begin{table}[h]
    \scriptsize
  \centering
  \caption{Comparison with existing few-shot object detection methods using nAP50 as evaluation metric on three PASCAL VOC Novel Split sets. Results are averaged over 10 random runs. $\dagger$ indicates that model is evaluated using the released code.}
  \begin{tabular}{l| c c c c c | c c c c c | c c c c c}
    \toprule
    \multirow{2}{*}{Method / Shots}  & \multicolumn{5}{c|}{Novel Split 1} & \multicolumn{5}{c|}{Novel Split 2} & \multicolumn{5}{c}{Novel Split 3} \\
    & 1 & 2 & 3 & 5 & 10 & 1 & 2 & 3 & 5 & 10& 1 & 2 & 3 & 5 & 10 \\
    \midrule
    FRCN+ft~\cite{yan2019meta}
    &9.9&15.6&21.6&28.0&35.6&9.4&13.8&17.4&21.9&29.8&8.1&13.9&19.0&23.9&31.0\\
    TFA w/fc~\cite{wang2020frustratingly}
    &22.9&34.5&40.4&46.7&52.0&16.9&26.4&30.5&34.6&39.7&15.7&27.2&34.7&40.8&44.6\\
    TFA w/cos~\cite{wang2020frustratingly} &25.3&36.4&42.1&47.9&52.8&18.3&27.5&30.9&34.1&39.5&17.9&27.2&34.3&40.8&45.6\\
    FsDetView~\cite{xiao2020few}
    &24.2&35.3&42.2&49.1&57.4&21.6&24.6&31.9&37.0&45.7&21.2&30.0&37.2&43.8&49.6\\
    TIP~\cite{li2021transformation} 
    &27.7&36.5&43.3&50.2&59.6&22.7&30.1&33.8&40.9&46.9&21.7&30.6&38.1&44.5&50.9\\
    DCNet~\cite{hu2021dense}&33.9&37.4&43.7&51.1&59.6&23.2&24.8&30.6&36.7&46.6&\textbf{32.3}&\textbf{34.9}&39.7&42.6&50.7\\
    FSCE~\cite{sun2021fsce} 
    &32.9&44.0&46.8&52.9&59.7&\textbf{23.7}&\textbf{30.6}&38.4&43.0&48.5&22.6&33.4&39.5&47.3&54.0\\
    \midrule
    TFA++$^{\dagger}$~\cite{sun2021fsce}
    &33.1&41.6&46.3&53.5&57.8&21.3&28.9&37.6&41.6&47.2&21.5&32.9&38.9&48.1&53.8\\
    Ours (KD-TFA++)
    &\textbf{35.4}&\textbf{46.2}&\textbf{48.1}&\textbf{56.5}&\textbf{60.7}&22.8&30.2&\textbf{39.2}&\textbf{44.0}&\textbf{48.9}&25.2&33.9&\textbf{41.3}&\textbf{50.7}&\textbf{55.9}\\
    \bottomrule
  \end{tabular}
  \label{tab:voc_multiple}
\end{table}

\section{Comparison with More Baseline Methods}
In Table~\ref{tab:morebaseline} we integrate our method into two more baselines: TFA w/ fc~\cite{wang2020frustratingly} and Retentive R-CNN~\cite{fan2021generalized}. It can be observed that our method consistently boosts the performance, which shows the effectiveness of our method.
\setlength\tabcolsep{6pt}
\begin{table}[h]
  \centering
  \caption{Performance of integrating our method into more baselines in terms of nAP50 on PASCAL VOC Novel split 1.}
    \scriptsize
  \begin{tabular}{l| c c c c c }
    \toprule
    \multirow{2}{*}{Methods / Shots} &  \multicolumn{5}{c}{nAP50} \\
      & 1 & 2 & 3 & 5 & 10 \\
    \midrule
    TFA w/ fc & 36.8 & 29.1 & 43.6 & 55.7 & 57.0 \\
    Ours (KD-TFA w/ fc) & \textbf{41.6} & \textbf{40.5} & \textbf{48.3} & \textbf{56.2} & \textbf{59.9}\\
    \midrule
    Retentive R-CNN & 42.4 & 45.8 & 45.9 & 53.7 & 56.1 \\
    Ours (KD-Retentive R-CNN) & \textbf{48.7} & \textbf{48.4} & \textbf{51.7} & \textbf{58.7} & \textbf{60.3}\\
    \bottomrule
  \end{tabular}
  \label{tab:morebaseline}
\end{table}

\section{More Ablation Studies}
\noindent\textbf{Effect of the number of visual words.}
Table~\ref{tab:num_words} shows the effect of the number of visual words. It can be observed that the performance first improves rapidly with the increase of visual words and then starts to degrade after 256. This is mainly resulted from the limited size of data corpus for learning the visual words.

\setlength\tabcolsep{6pt}
\begin{table}[ht]
\centering
\scriptsize
\caption{Effect of the number of visual words.}
\begin{tabular}{c| c c c }
    \toprule
    \multirow{2}{*}{Number} &  \multicolumn{3}{c}{nAP50} \\
      & 3 & 5 & 10 \\
    \midrule
    64& 50.0 & 57.2 & 61.3 \\
    128&51.6 & 61.0 & 62.3\\
    256& \textbf{52.5} & \textbf{62.1} & \textbf{64.2} \\
    512& 51.6 & 59.2 & 62.3 \\
    \bottomrule
  \end{tabular}
  \label{tab:num_words}
\end{table}

\noindent\textbf{Knowledge distillation vs initialization of the object detector vs multi-task learning.} We further explore other methods to learn a generalizable detector. As shown in Row 2 of Table~\ref{tab:method}, using the backbone pre-trained by PPC~\cite{xie2021propagate} to initialize the detector yields little improvement over the baseline. Row 3 shows that the performance degrades when using PPC for multi-task learning, presumably because PPC aims to distinguish between pixels, which is not entirely consistent with the objective of object detection. 

\setlength\tabcolsep{6pt}
\begin{table}[ht]
    \centering
    \scriptsize
    \caption{Performance of different ways of using PPC on VOC Novel Split 1.}
    \begin{tabular}{l | c  c  c}
        \toprule
        \multirow{2}{*}{Methods / Shots} &  \multicolumn{3}{c}{nAP50} \\
      & 3 & 5 & 10 \\
        \midrule
        Baseline & 47.2 & 57.2 & 60.8\\
        Initialization & 46.4 & 57.3 & 61.2\\
        Multi-task Learning & 45.9 & 55.1 & 60.3 \\
        Knowledge Distillation (Ours) & \textbf{52.5} & \textbf{62.1} & \textbf{64.2} \\
        \bottomrule
    \end{tabular}
    \label{tab:method}
\end{table}

\section{More Qualitative Detection Results}
\begin{figure}
  \centering
   \includegraphics[width=1.0\linewidth]{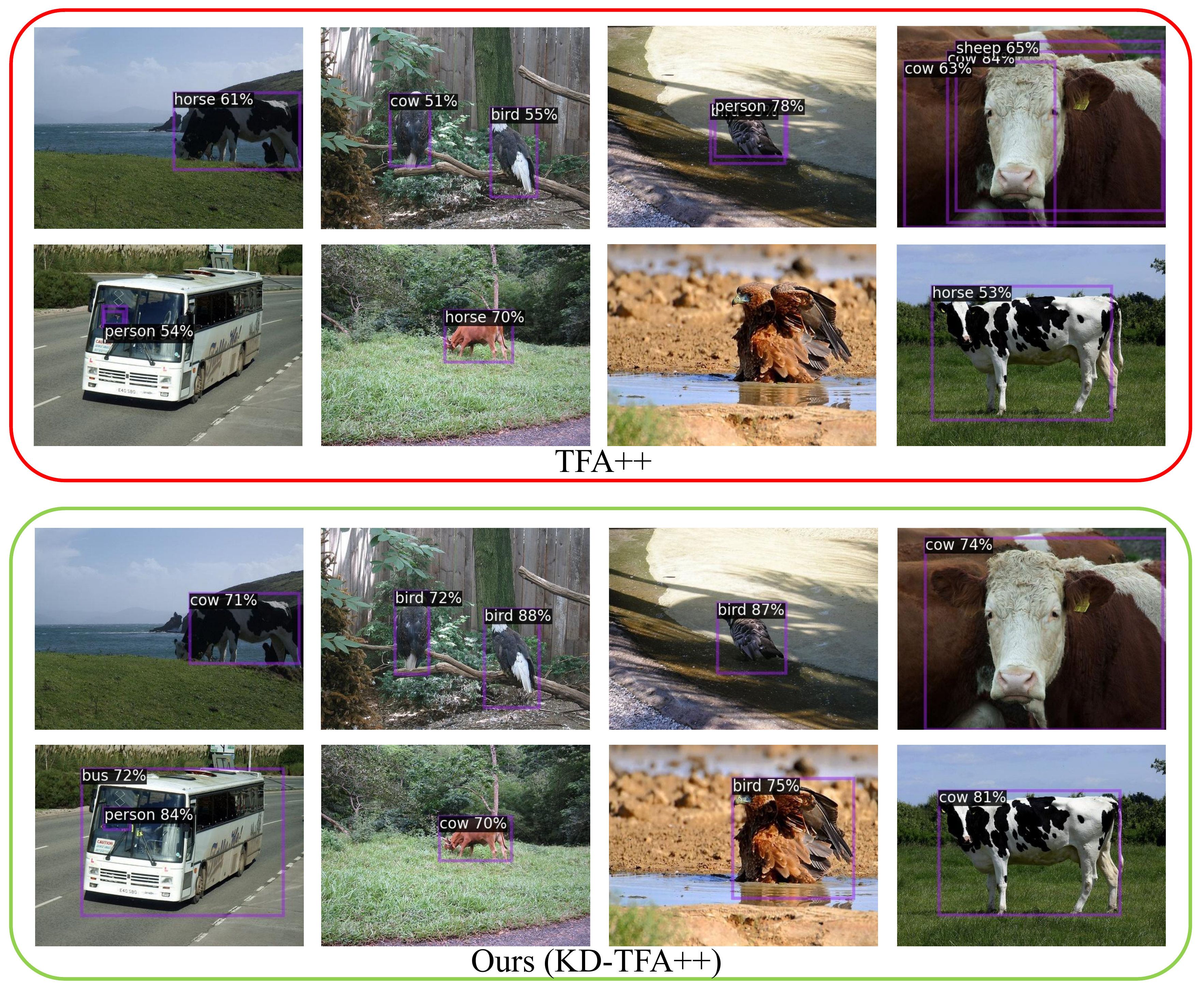}
   \caption{Detection results of TFA++~\cite{sun2021fsce} and our method under PASCAL VOC Novel Split1 10-shot setting.} 
   \label{fig:more_vis}
\end{figure}

We provide more qualitative detection results under 10-shot setting of PASCAL VOC Novel Split1. As shown in Figure~\ref{fig:more_vis}, our method reduces 
the appearance of each type of errors such as missing detections and misclassifying novel objects.

\end{document}